\documentclass[runningheads]{llncs}
\usepackage{graphicx}
\usepackage{amsmath,amssymb} 
\usepackage{color}

\graphicspath{ {figs/} }

\usepackage{booktabs}
\usepackage{makecell}
\usepackage{tabu,multirow}
\usepackage{pbox}
\usepackage{multicol}
\usepackage{hhline}
\usepackage{subfigure}

\usepackage[dvipsnames]{xcolor}

\newcommand{\oursshort}[0]{MF\xspace}
\newcommand{\ours}[0]{MF-Net\xspace}
\newcommand{\ourslong}[0]{multi-fiber\xspace}
\newcommand{\ourslongcaps}[0]{Multi-Fiber\xspace}
\newcommand{\ourslongcap}[0]{Multi-fiber\xspace}
\newcommand{\switch}[0]{multiplexer\xspace}

\usepackage{xspace}
\makeatletter
\DeclareRobustCommand\onedot{\futurelet\@let@token\@onedot}
\def\@onedot{\ifx\@let@token.\else.\null\fi\xspace}
 
\def\ie{\emph{i.e}\onedot}

\makeatother

\begin{document}
\title{Multi-Fiber Networks for Video Recognition}

\titlerunning{Multi-Fiber Networks}
%
\author{Yunpeng~Chen\inst{1} \and
Yannis~Kalantidis\inst{2} \and
Jianshu~Li\inst{1} \and \\
Shuicheng~Yan\inst{3,1} \and
Jiashi~Feng\inst{1}}
%
\authorrunning{Y. Chen, Y. Kalantidis, J. Li, S. Yan and J. Feng}
%

\institute{National University of Singapore \\
\and
Facebook Research\\
\and
Qihoo 360 AI Institute\\
\email{\{chenyunpeng,~jianshu\}@u.nus.edu, yannisk@fb.com, \{eleyans,~elefjia\}@nus.edu.sg}}
\maketitle              

\begin{abstract}
In this paper, we aim to reduce the computational cost of spatio-temporal deep neural networks, making them run as fast as their 2D counterparts while preserving state-of-the-art accuracy on video recognition benchmarks. To this end, we present the  novel \textit{\ourslongcaps} architecture that slices a complex neural network into an ensemble of lightweight networks or \textit{fibers} that run through the network. To facilitate information flow between fibers we further incorporate \switch modules and end up with an architecture that reduces the computational cost of 3D networks by an order of magnitude, while increasing recognition performance at the same time. Extensive experimental results show that our \ourslong architecture significantly boosts the efficiency of existing convolution networks for both image and video recognition tasks, achieving state-of-the-art performance on UCF-101, HMDB-51 and Kinetics datasets. 
Our proposed model requires over {$9 \times$} and {$13 \times$} less computations than the I3D~\cite{carreira2017quo} and R(2+1)D~\cite{tran2017closer} models, respectively, yet providing higher accuracy.

\end{abstract}

\section{Introduction}
\label{sec:intro}

With the aid of deep convolutional neural networks, image understanding has achieved remarkable success in the past few years. Notable examples include residual networks~\cite{he2016deep} for image classification, FastRCNN~\cite{girshick2015fast} for object detection, and Deeplab~\cite{chen2016deeplab} for semantic segmentation, to name a few. However, the progress of deep neural networks for video analysis still lags their image counterparts, mostly due to the extra computational cost and complexity of spatio-temporal inputs.

The temporal dimension of videos contains valuable motion information that needs to be incorporated for video recognition tasks.
A popular and effective way of reasoning spatio-temporally is to use spatio-temporal or 3D convolutions~\cite{karpathy2014large,tran2015learning} in deep neural network architectures to learn video representations.
A 3D convolution is an extension of the 2D (spatial) convolution, which has three-dimensional kernels that also convolve along the temporal dimension. The 3D convolution kernels can be used to build 3D CNNs (Convolutional Neural Networks) by simply replacing the 2D spatial convolution kernels. This keeps the model end-to-end trainable. State-of-the-art video understanding models, such as Res3D~\cite{tran2015learning} and I3D~\cite{carreira2017quo} build their CNN models in this straightforward manner. They use multiple layers of 3D convolutions to learn robust video representations and achieve top accuracy on multiple datasets, albeit with high computational overheads. Although recent approaches use decomposed 3D convolutions~\cite{tran2017closer,xie2017rethinking} or group convolutions~\cite{hara2018can} to reduce the computational cost, the use of spatio-temporal models still remains  prohibitive for practical large-scale applications. 
For example, regular 2D CNNs require around 10s GFLOPs for processing a single frame, while 3D CNNs currently require more than 100 GFLOPs for a single clip\footnote{\emph{E.g.} the popular ResNet-152~\cite{he2016deep} and VGG-16~\cite{simonyan2014very} models require 11 GFLOPs and 15 GFLOPs, respectively, for processing a frame, while I3D~\cite{carreira2017quo} and R(2+1)D-34~\cite{tran2017closer}  require 108 GFLOPs and 152 GFLOPs, respectively.}. 
\emph{We argue that a clip-based model should be able to highly outperform frame-based models at video recognition tasks for the same computational cost, given that it has the added capacity of reasoning spatio-temporally.}
\nocite{zheng_cvpr16_scnn,zheng_cvpr17_cdc}

In this work, we aim to substantially improve the efficiency of 3D CNNs while preserving their state-of-the-art accuracy on video recognition tasks. Instead of decomposing the 3D convolution filters as in~\cite{tran2017closer,xie2017rethinking}, we focus on the other source of computational overhead for 3D CNNs, the large input tensors. We propose a sparsely connected architecture, the \textit{\ourslongcaps} network, where each unit in the architecture is essentially composed of multiple \textit{fibers}, \ie lightweight 3D convolutional networks that are independent from each other as shown in Fig~\ref{fig_multibranch_unit}(c). The overall network is thus sparsely connected and the computational cost is reduced by approximately $N$ times, where $N$ is the number of fibers used. To improve information flow across fibers, we further propose a lightweight \switch module, that redirects information between parallel fibers if needed and is attached at the head of each residual block. This way, with a minimal computational overhead, representations can be shared among multiple fibers, and the overall capacity of the model is increased. 

Our main contributions can be summarized as follows:

1) We propose a highly efficient \ourslong architecture, verify its effectiveness by evaluating it 2D convolutional neural networks for image recognition and show that it can boost performance when embedded on common compact models.

2) We extend the proposed architecture to spatio-temporal convolutional networks and propose the \ourslongcaps network (\ours) for learning robust video representations with significantly reduced computational cost, \ie about an order of magnitude less than the current state-of-the-art 3D models.

3) We evaluate our \ourslong network on multiple video recognition benchmarks and outperform recent related methods with several times lower computational cost on the Kinetics, UCF-101 and HMDB51 datasets.

\section{Related Work}
\label{sec:related}

When it comes to video models, the most successful approaches utilize deep learning and can be split into two major categories: models based on spatial or 2D convolutions and those that incorporate spatio-temporal or 3D convolutions.

The major advantage of adopting 2D CNN based methods is their computational efficiency. One of the most successful approaches in this category is the Two-stream Network~\cite{simonyan2014two} architecture. It is composed of two 2D CNNs, one working on frames and another on optical flow. Features from the two modalities are fused at the final stage and achieved high video recognition accuracy. Multiple approaches have extended or incorporated the two-stream model~\cite{feichtenhofer2016convolutional,ng2015beyond,wang2017appearance,tran2017two} and since they are built on 2D CNNs are very efficient, usually requiring less than 10 GFLOPS per frame. In a very interesting recent approach, CoViAR~\cite{wu2017compressed} further reduces computations to 4.2 GFLOPs per frame in average, by directly using the motion information from compressed frames and sharing motion features across frames. However, as these approaches rely on pre-computed motion features to capture temporal dependencies, they usually perform worse than 3D convolutional networks, especially when large video datasets are available for pre-training, such as Sports-1M~\cite{KarpathyCVPR14} and Kinetics~\cite{kay2017kinetics}.
\nocite{zheng_eccv18_autoloc,zheng_eccv18_oad}

On the contrary, 3D convolution neural networks are naturally able to learn motion features from raw video frames in an end-to-end manner. Since they use 3D convolution kernels that model both spatial and temporal information, rather than 2D kernels which just model spatial information, more complex relations between motion and appearance can be learned and captured. C3D~\cite{tran2015learning} is one of the early methods successfully applied to learning robust video features. It builds a VGG~\cite{simonyan2014very} alike structure but uses $3\times3\times3$ kernels to capture motion information. The Res3D~\cite{tran2017convnet} makes one step further by taking the advantage of residual connections to ease the learning process. Similarly, I3D~\cite{carreira2017quo} proposes to use the Inception Network~\cite{szegedy2015going} as the backbone network rather than residual networks to learn video representations. However, all of the methods suffer from high computational cost compared with regular 2D CNNs due to the newly added temporal dimension. Recently, S3D~\cite{xie2017rethinking} and R(2+1)D~\cite{tran2017closer} are proposed to use one $1\times3\times3$ convolution layer followed by another $3\times1\times1$ convolutional layer to approximate a full-rank 3D kernel to reduce the computations of a full-rank $3\times3\times3$ convolutional layer while achieving better precision. However, these methods still suffer from an order of magnitude more computational cost than their 2D competitors, which makes it difficult to train and deploy them in practical applications.

The idea of using spare connections to reduce the computational cost is similar to low-power networks built for mobile devices~\cite{howard2017mobilenets,sandler2018inverted,zhang2017shufflenet} as well as other recent approaches that try to sparsify parts of the network either through group convolutions~\cite{xie2017aggregated} or through learning connectivity~\cite{ahmed2018maskconnect}. However, our proposed network is built for solving video recognition tasks and proposed different strategies that can also benefit existing low-power models, \emph{e.g.} MobileNet-v2 \cite{sandler2018inverted}. We further discuss the differences of our architecture and compare against the most related and state-of-the-art methods in Sections 3 and 4.

\section{\ourslongcaps Networks}
\label{sec:method}

The success of models that utilize spatio-temporal convolutions~\cite{tran2015learning,carreira2017quo,tran2017closer,xie2017rethinking,hara2018can} suggests that it is crucial to have kernels spanning both the spatial and temporal dimensions. Spatio-temporal reasoning, however, comes at a cost: Both the convolutional kernels and the input-output tensors are multiple times larger.

In this section, we start by describing the basic module of our proposed model, \emph{i.e.}, the \ourslong unit. This unit can effectively reduce the number of connections within the network and enhance the model efficiency. It is generic and compatible with both 2D and 3D CNNs. For clearer illustration, we first demonstrate its effectiveness by embedding it into 2D convolutional architectures and evaluating its efficiency benefits for image recognition tasks. We then introduce its spatio-temporal 3D counterpart and discuss specific design choices for video recognition tasks.

\subsection{The \ourslongcap Unit}

\begin{figure}[t]	
	\center
	\resizebox{1.0\textwidth}{!}{
		\includegraphics[width=\textwidth]{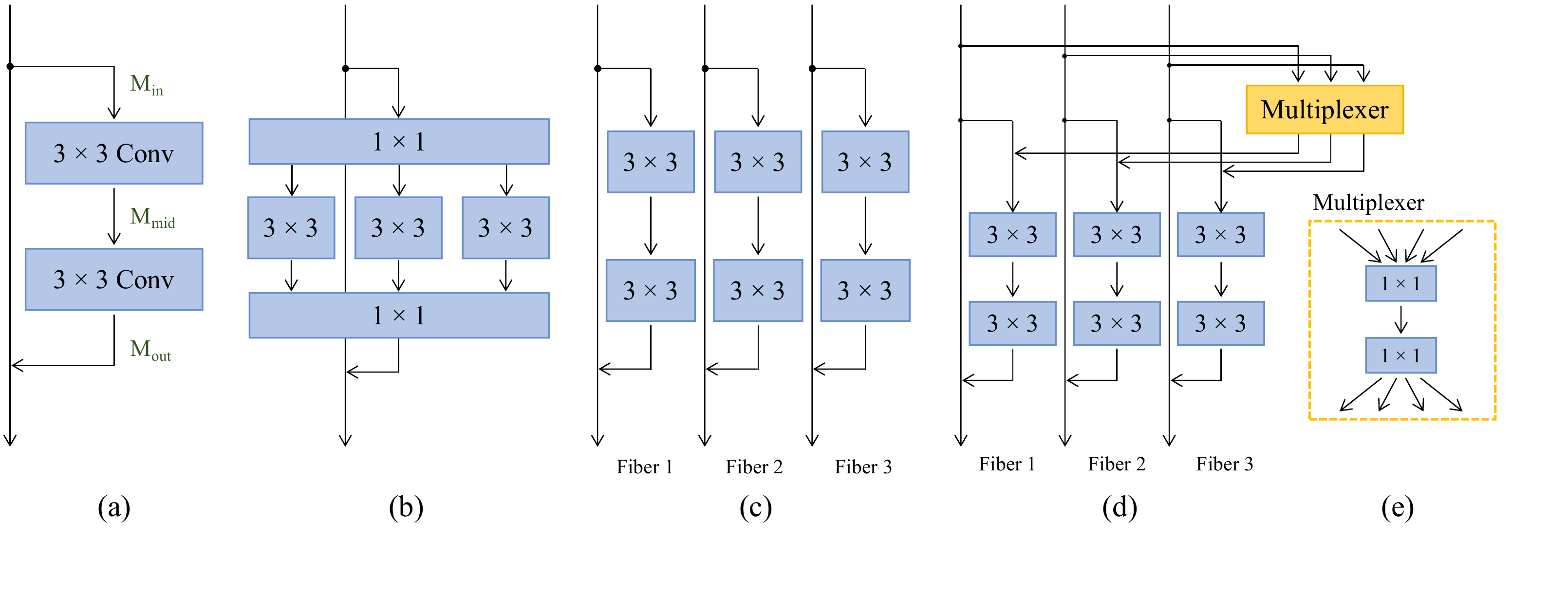}
	}
	\caption{From ResNet to \ourslong.
    (a) A residual unit with two $3\times3$ convolution layers. (b) Conventional Multi-Path design, \emph{e.g.} ResNeXt~\cite{xie2017aggregated}. (c) The proposed \ourslong design consisting of multiple separated lightweight residual units, called fibers. (d) The proposed \ourslong architecture with a \textit{\switch} for transferring information across separated fibers. (e) The architecture details of a \switch. It consists of two linear projection layers, one for dimension reduction and the other for dimension expansion.}
	\label{fig_multibranch_unit}
\end{figure}

The proposed \ourslong unit is based on the highly modularized residual unit~\cite{he2016deep}, which is easy to train and deploy. As shown in Figure~\ref{fig_multibranch_unit}(a), the conventional residual unit uses two convolutional layers to learn features, which is straightforward but computationally expensive. 
To see this, let $M_{in}$ denote the number of input channels,  $M_{mid}$ denote the number of middle channels, and $M_{out}$ denote the number of output channels. Then the total number of connections between these two layers can be computed as
\begin{equation}
\label{eq_3x3residual_org_unit_cost}
\text{\# Connections} = M_{in} \times M_{mid} + M_{mid} \times M_{out}.
\end{equation}
For simplicity, we ignore the dimensions of the input feature maps and convolution kernels which are constant. Eqn.~\eqref{eq_3x3residual_org_unit_cost} indicates that the number of connections is quadratic to the width of the network, thus increasing  the width of the unit by a factor of $k$ would result in $k^2$ times more computational cost.

To reduce the number of connections that are essential to the overall computation cost, we propose to \emph{slice} the complex residual unit into $N$ parallel and separated paths (called \emph{fibers}), each of which is isolated from the others, as shown in Figure~\ref{fig_multibranch_unit}(c). In this way, the overall width of the unit remains the same, but the number of connections is reduced by a factor of $N$:
\begin{align}
\label{eq_3x3residual_unit_cost}
\text{\# Connections} & = N \times (M_{in}/N \times M_{mid}/N + M_{mid}/N \times M_{out}/N) \nonumber
\\
						     & = (M_{in} \times M_{mid} + M_{mid} \times M_{out})/N.
\end{align}
We set $N=16$ for all our experiments, unless otherwise stated.
As we show experimentally in the following section, such a slicing strategy is intuitively simple yet effective. At the same time, however, slicing isolates each path from the others and blocks any information flow across them. This may result in limited learning capacity for data representations since one path cannot access and utilize the feature learned from the others. 
In order to recover part of the learning capacity, recent approaches that partially use slicing like ResNeXt~\cite{xie2017aggregated}, Xception~\cite{chollet2017xception} and MobileNet~\cite{howard2017mobilenets,sandler2018inverted} choose to only slice a small portion of layers and still use fully connected parts. The majority of layers ($>60\%$) remains unsliced and dominates the computational cost, becoming the efficiency bottleneck. 
ResNeXt~\cite{xie2017aggregated}, for example, uses fully connected convolution layers at the beginning and end of each unit, and only slices the second layer as shown on Figure~\ref{fig_multibranch_unit}(b). However, these unsliced layers dominate the computation cost and become the bottleneck.
Different from only slicing a small portion of layers, we propose to slice the entire residual unit creating multiple fibers. To facilitate information flow, we further attach a lightweight bottleneck component we call the \emph{\switch} that operates across fibers, in a residual manner.

The \switch acts as a router that redirects and amplifies features from all fibers. As shown in Figure~\ref{fig_multibranch_unit}(e), the \switch first gathers features from all fibers using a $1\times1$ convolution layer, and then redirects them to specific fibers using the following $1\times1$ convolution layer. The reason for using two $1\times1$ layers instead of just one is to lower the computational overhead: we set the number of the first-layer output channels to be $k$ times smaller than its input channels, so that the total cost would be reduced by a factor of $k/2$ compared with using a single $1\times1$ layer. The parameters within the \switch are randomly initialized and automatically adjusted by back-propagation end-to-end to maximize the performance gain for the given task. Batch normalization and ReLU nonlinearities are used before each layer. 
Figure~\ref{fig_multibranch_unit}(d) shows the full \ourslong network, where the proposed \switch is attached at the beginning of the multi-fiber unit for routing features extracted from other paralleled fibers. 

We note that, although the proposed \ourslong architecture is motivated to reduce the number of connections for 3D CNNs to alleviate high computational cost, it is also applicable to 2D CNNs to further enhance  efficiency of existing 2D architectures. To demonstrate this and verify  effectiveness of the proposed architecture, we conduct several  studies on 2D image classification tasks at first.

\subsection{Justification of the \ourslongcap Architecture}

We experimentally study the effectiveness of the proposed \ourslong architecture by applying it on 2D CNNs for image classification and the ImageNet-1k dataset~\cite{krizhevsky2012imagenet}. We use one of the most popular 2D CNN model, residual network (ResNet-18)~\cite{he2016deep}, and  the most computationally efficient ModelNet-v2~\cite{sandler2018inverted} as the backbone CNN in the following studies.

Our implementation is based on the code released by \cite{chen2017dual} using MXNet~\cite{chen2015mxnet} on a cluster of 32 GPUs. The initial learning rate is set to $0.5$ and decreases exponentially. We use a batch size of 1,024 and train the network for 360,000 iterations. As suggested by prior work~\cite{howard2017mobilenets}, we use less data augmentations for obtaining better results. Since the above training strategy is different from the one used in our baseline methods~\cite{he2016deep,sandler2018inverted}, we report both our reproduced results and the reported results in their papers for fair comparison.

\begin{figure}
\centering
\subfigure[ResNet-18]{\includegraphics[width=5cm]{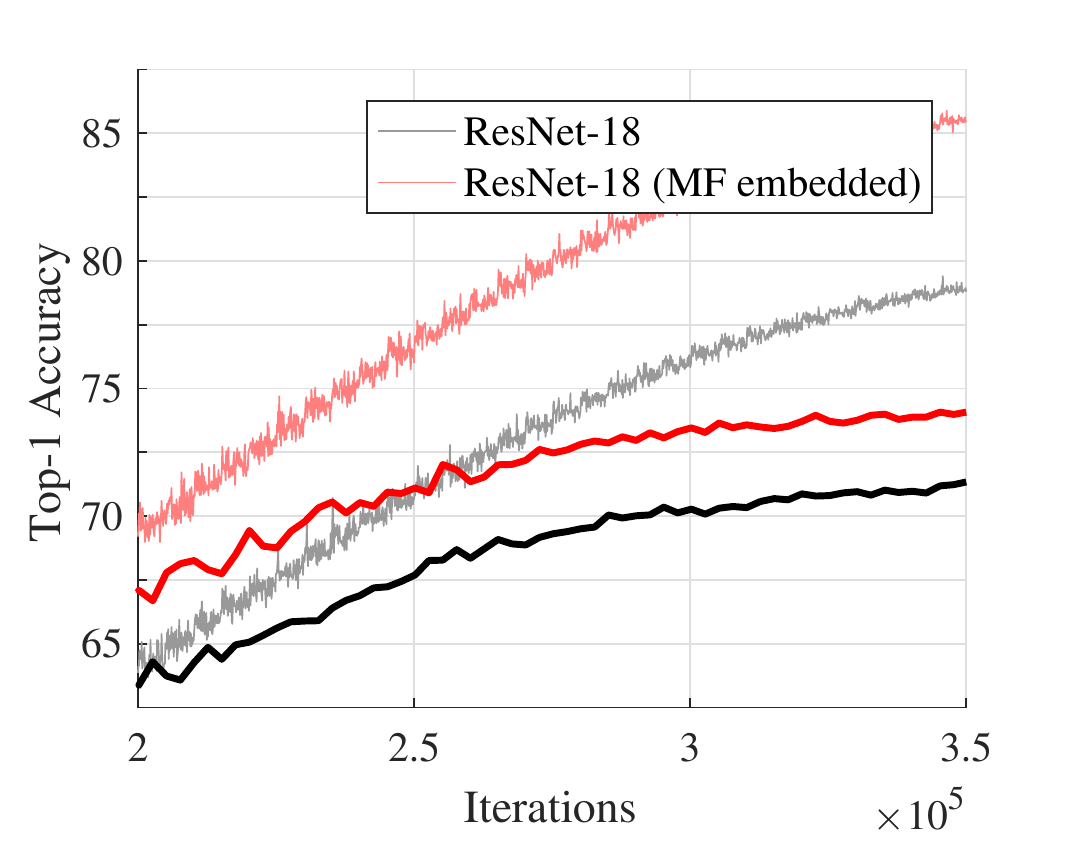}}
\subfigure[MobileNet-v2]{\includegraphics[width=5cm]{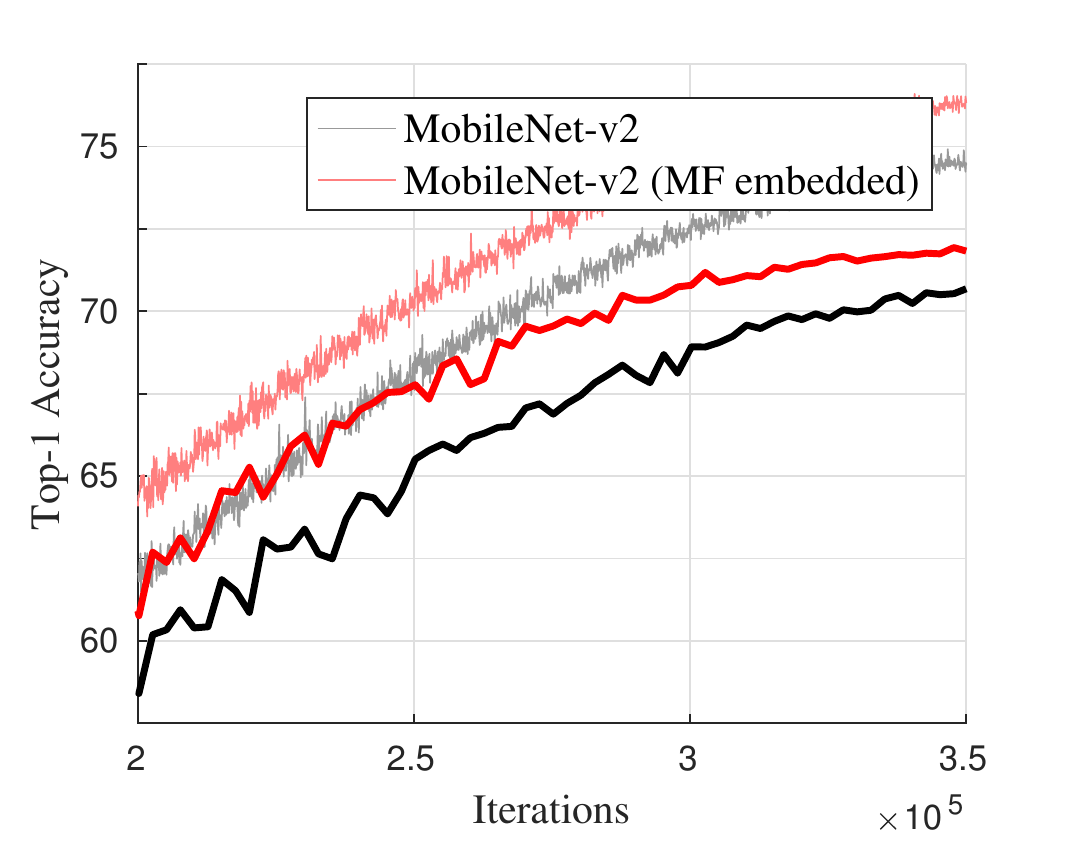}}
\caption{Training and validation accuracy on the ImagaNet-1k dataset for (a) ResNet-18 and (b) MobileNet-v2 backbones respectively. The red lines stand for performance of the model with our proposed \ourslong unit. The black lines show performance of our reproduced baseline model using exactly the same training settings as our method. The line thickness indicates results on the validation set (the ticker one) or the training set (the thinner one).}
\label{fig_imnet_curve}
\end{figure}
The training curves in Figure~\ref{fig_imnet_curve} plot the training and validation accuracy on ImageNet-1k during the last several  iterations. One can observe that the network with our proposed Multi-fiber (MF) unit can consistently achieve higher training and validation accuracy than the baseline
models, with the same number of iterations.  Moreover, the resulted model has a smaller number of parameters and is more efficient (see Table~\ref{imgnet}).  This demonstrates that  embedding the proposed MF unit indeed helps reduce the model redundancy, accelerates the learning process and improves the overall model generalization ability. Considering the final training accuracy of the ``MF embedded'' network is significantly higher than the baseline networks and all the network models adopt the same regularization settings, the MF units are also demonstrated to be able to improve the learning capacity of the baseline networks.

\begin{table}[t]
\caption{Efficiency comparison on the ImageNet-1k validation set. ``\oursshort'' stands for ``\ourslong unit'', and Top-1/Top-5 accuracies are evaluated on a $224\times224$ single center crop \cite{he2016deep}. ``\ours'' is our proposed network, with the architecture shown in ~\ref{MBNET-arch}. The ResNeXt row presents results for a ResNeXt-26 model of our design that has about the same number of FLOPS as \ours.} 
\centering
\resizebox{1.\textwidth}{!}{
  \begin{tabular}{>{\centering}p{8cm}|>{\centering}p{2cm}|>{\centering}p{2cm}|>{\centering}p{2cm}|c}
  \toprule    
  Model                       					 & Top-1 Acc.   & Top-5 Acc    & \#Params   &~~~~~FLOPs~~~~\\
  \midrule    
  ResNet-18~\cite{he2016deep} 					 &  69.6 \%     &  89.2 \%     &  11.7 M    &  1.8 G    \\
  ResNet-18 (reproduced)						 &  71.4 \%     &  90.2 \%     &  11.7 M    &  1.8 G    \\
  ResNet-18 (\oursshort embedded)     			 &  74.3 \%     &  92.1 \%     &   9.6 M    &  1.6 G    \\
  \midrule    
  ResNeXt-26 ($8 \times 16d$)			 		 &  72.8 \%		&  91.1 \%     &   6.3 M    &  1.1 G    \\
  ResNet-50~\cite{he2016deep}    				 &  75.3 \%     &  92.2 \%     &  25.5 M    &  4.1 G    \\
  \midrule    
  MobileNet-v2 (1.4)~\cite{sandler2018inverted}  &  74.7 \%     &   --         &   6.9 M    &  585 M    \\
  MobileNet-v2 (1.4) (reproduced)				 &  72.2 \%     &  90.8 \%     &   6.9 M    &  585 M    \\
  MobileNet-v2 (1.4) (\oursshort embedded) 	     &  73.0 \%     &  91.1 \%     &   6.0 M    &  578 M    \\
  \midrule    
  \ours ($N = 12$)			     				 &  74.5 \%     &  92.0 \%     &   5.9 M    &  895 M    \\
  \ours ($N = 16$)  			 				 &  74.6 \%     &  92.0 \%     &   5.8 M    &  861 M    \\
  \ours ($N = 24$)  			 				 &  75.4 \%     &  92.5 \%     &   5.8 M    &  897 M    \\
  \midrule    
  \ours ($N = 16$, w/o \switch)	     		   	 &  70.2 \%     &  89.4 \%     &   4.5 M    &  600 M    \\
  \ours ($N = 16$, w/o \switch, deeper \& wider) &  71.0 \%     &  90.0 \%     &   6.4 M    &  897 M    \\
  \bottomrule 
  \end{tabular}
}
\label{imgnet}
\end{table}
Table~\ref{imgnet} presents results on the validation set for Imagenet-1k. By simply replacing the original residual unit with our proposed \ourslong one, we improve the Top-1/Top-5 accuracy by $2.9$\%/$1.9$\% upon ResNet-18 with smaller model size (9.6M vs. 11.7M ) and lower FLOPs (1.6G vs. 1.8G). The performance gain also stands for the more efficient low-complexity MobileNet-v2: introducing the \ourslong unit also boosts its Top-1/Top-5 accuracy by $0.8$\%/$0.3$\% with smaller model size (6.0M vs. 6.9M) and lower FLOPs (578M vs. 585M), clearly demonstrating its effectiveness. 
We note that our reproduced MobileNet-v2 has slightly lower accuracy than the reported one in~\cite{sandler2018inverted} due to difference in the batch size, learning rate and update policy. But with the same training strategy, our reproduced ResNet-18 is $1.8$\% better than the reported one~\cite{he2016deep}.

The two bottom sections of Table~\ref{imgnet} further show ablation studies of our \ours, with respect to the number of fibers $N$ and with/without the use of the \switch. As we see, increasing the number of fibers increases performance, while performance drops significantly when removing the \switch unit, demonstrating the importance of sharing information between fibers. Overall, we see that our 2D \ourslong network can perform as well as the much larger ResNet-50~\cite{he2016deep}, that has $25.5$M parameters and requires 4.1 GFLOPS\footnote{It is worth noting that in terms of wall-clock time measured on our server, our \ours is only slightly (about 30\%) faster than the highly optimized implementation of ResNet-50. We attribute this to the unoptimized implementation of group convolutions in CuDNN and foresee faster actual running times in the near future when group convolution computations are well optimized.}.

\subsection{Spatio-temporal \ourslongcap Networks}
\label{Network_Architecture}

\begin{figure}[t]	
	\center
	\resizebox{1.0\textwidth}{!}{	
		\includegraphics{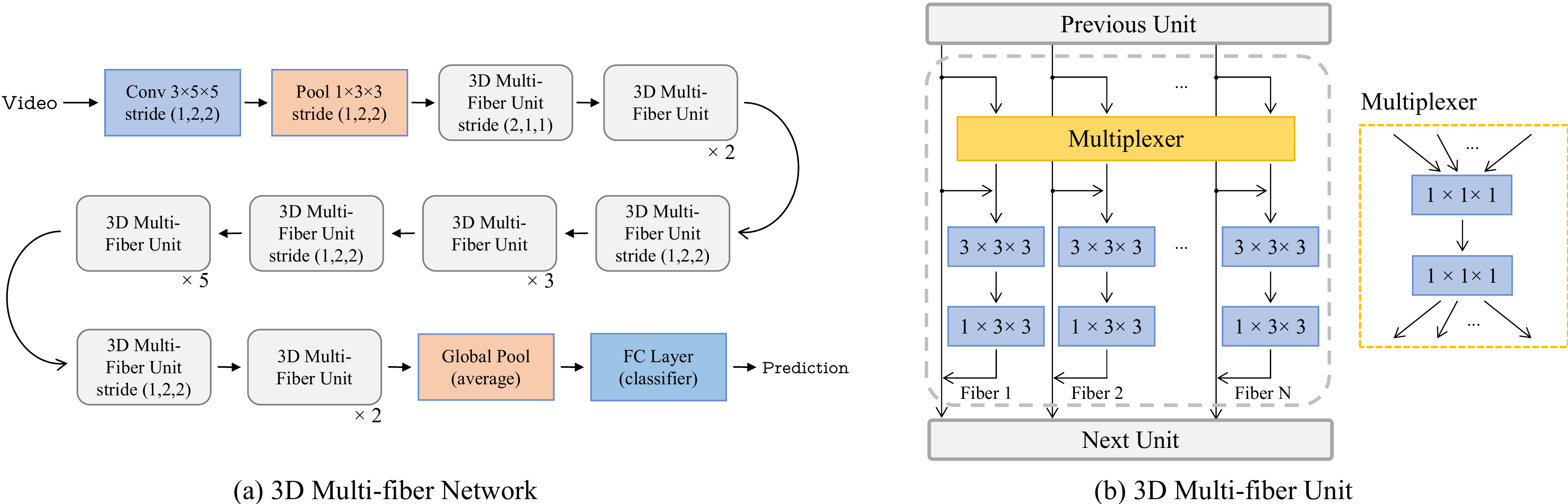}
	}
	\caption{Architecture of 3D \ourslong network. (a) The overall architecture of 3D \ourslongcap Network. (b) The internal structure of each \ourslongcap Unit. Note that only the first $3\times3$ convolution layer has expanded on the 3rd temporal dimension for lower computational cost.}
	\label{fig_network_3d}
\end{figure}

\begin{table}[t]
\centering
\caption{\ourslongcap Network architecture. The ``2D \ours'' takes images as input, while the ``3D \ours'' takes frames, \emph{i.e.} video clips, as input. Note, the complexity is evaluated with FLOPs, \emph{i.e.} floating-point multiplication-adds. The stride of ``3D \ours'' is denoted by ``(temporal stride, height stride, width stride)'', and the stride of ``2D \ours'' is denoted by ``(height stride, width stride)''.}
\resizebox{1.\textwidth}{!}{
  \begin{tabular}{>{\centering}p{2.5cm}||>{\centering}p{1.5cm}|>{\centering}p{2cm}||>{\centering}p{2.5cm}|>{\centering}p{1.5cm}||>{\centering}p{2.5cm}|c}
  \toprule
  \multirow{2}{*}{layer} & \multirow{2}{*}{Repeat} & \multirow{2}{*}{\#Channel} & \multicolumn{2}{c||}{2D \ours} & \multicolumn{2}{c}{3D \ours} \\
  \cline{4-7}
                         &      &                       & Output Size                   & Stride & Output Size              				&~~~~Stride~~~~\\
  \hhline{=======}
  Input                  &      & 3                     & $224\times224$                &        & $16\times224\times224$   				&        \\
  \hline
  Conv1                  & 1    & \multirow{2}{*}{16}   & $112\times112$                & (2,2)  & $16\times112\times112$   				& (1,2,2)  \\
  MaxPool                &      &                       &   $56\times56$                & (2,2)  &   $16\times56\times56$   				& (1,2,2)  \\
  \hline
  \multirow{2}{*}{Conv2} & 1    & \multirow{2}{*}{96}   & \multirow{2}{*}{$56\times56$} & (1,1)  & \multirow{2}{*}{$8\times56\times56$}	& (2,1,1)  \\
                         & 2    &                       &                               & (1,1)  &                          				& (1,1,1)  \\
  \hline
  \multirow{2}{*}{Conv3} & 1    & \multirow{2}{*}{192}  & \multirow{2}{*}{$28\times28$} & (2,2)  & \multirow{2}{*}{$8\times28\times28$}	& (1,2,2)  \\
                         & 3    &                       &                               & (1,1)  &                          				& (1,1,1)  \\
  \hline
  \multirow{2}{*}{Conv4} & 1    & \multirow{2}{*}{384}  & \multirow{2}{*}{$14\times14$} & (2,2)  & \multirow{2}{*}{$8\times14\times14$}	& (1,2,2)  \\
                         & 5    &                       &                               & (1,1)  &                          				& (1,1,1)  \\
  \hline
  \multirow{2}{*}{Conv5} & 1    & \multirow{2}{*}{768}  & \multirow{2}{*}{$7\times7$}   & (2,2)  & \multirow{2}{*}{$8\times7\times7$}  	& (1,2,2)  \\
                         & 2    &                       &                               & (1,1)  &                          				& (1,1,1)  \\
  \hline
  AvgPooling             &      &                       & $1\times1$                    &        & $1\times1\times1$        				&        \\
  \hline
  FC                     &      &                       & 1000                          &        & 400                      				&        \\
  \hhline{=======}
  \#Params               &      &                       & \multicolumn{2}{c||}{5.8 M}            & \multicolumn{2}{c}{8.0 M}                      \\
  \hline
  FLOPs                  &      &                       & \multicolumn{2}{c||}{861 M}            & \multicolumn{2}{c}{11.1 G}                     \\
  \bottomrule
  \end{tabular}
}
\label{MBNET-arch}
\end{table}

In this subsection, we extend out \ourslong architecture to spatio-temporal inputs and present a new architecture for 3D convolutional networks and video recognition tasks.
The design of our spatio-temporal \ourslong network follows that of the ``ResNet-34''~\cite{he2016deep} model, with a slightly different number of channels for lower GPU memory cost on processing videos. In particular, we reduce the number of channels in the first convolution layer, \emph{i.e.} ``Conv1'', and increase the number of channels in the following layers, \emph{i.e.} ``Conv2-5'', as shown in Table~\ref{MBNET-arch}.
This is because the feature maps in the first several layers have high resolutions and consume exponentially more GPU memory than the following layers for both training and testing.

The detailed network design is shown in Table~\ref{MBNET-arch}, where we first design a 2D \ours and then ``inflate''~\cite{carreira2017quo} its 2D convolutional kernels to 3D ones to build the 3D \ours. The 2D \ours is used as a pre-trained model for initializing the 3D \ours. Several recent works advocate separable convolution which uses two separate layers to replace one $3\times3$ layer \cite{tran2017closer,xie2017rethinking}. Even though it may further reduce the computational cost and increase the accuracy, we do not use the separable convolution due to its high GPU memory consumption, considering video recognition application.

Figure~\ref{fig_network_3d} shows the inner structure of each 3D \ourslong unit after the ``inflation'' from 2D to 3D. We note that all convolutional layers use 3D convolutions thus the input and output features contain an additional temporal dimension for preserving motion information.

\section{Experiments}
\label{sec:experiments}

We evaluate the proposed \ourslong network on three benchmark datasets, Kinetics~\cite{kay2017kinetics}, UCF-101~\cite{soomro2012ucf101} and HMDB51~\cite{kuehne2011hmdb}, and compare the results with other state-of-the-art models. All experiments are conducted using PyTorch~\cite{paszke2017pytorch} with input size of $16\times224\times224$ for both training and testing. Here $16$ is the number of frames for each input clip. During testing, videos are resized to resolution $256\times256$, and we average clip predictions randomly sampled from the long video sequence to obtain the video predictions.

\subsection{Video Classification with Motion Trained from Scratch}

In this subsection, we study the effectiveness of the proposed model on learning video representations when motion features are trained from scratch. We use the large-scale Kinetics~\cite{kay2017kinetics} benchmark dataset for evaluation, which consists of approximately $300,000$ videos from $400$ action categories.

\begin{figure}[t]
\centering
\subfigure[]{\resizebox{0.4\textwidth}{!}{\includegraphics{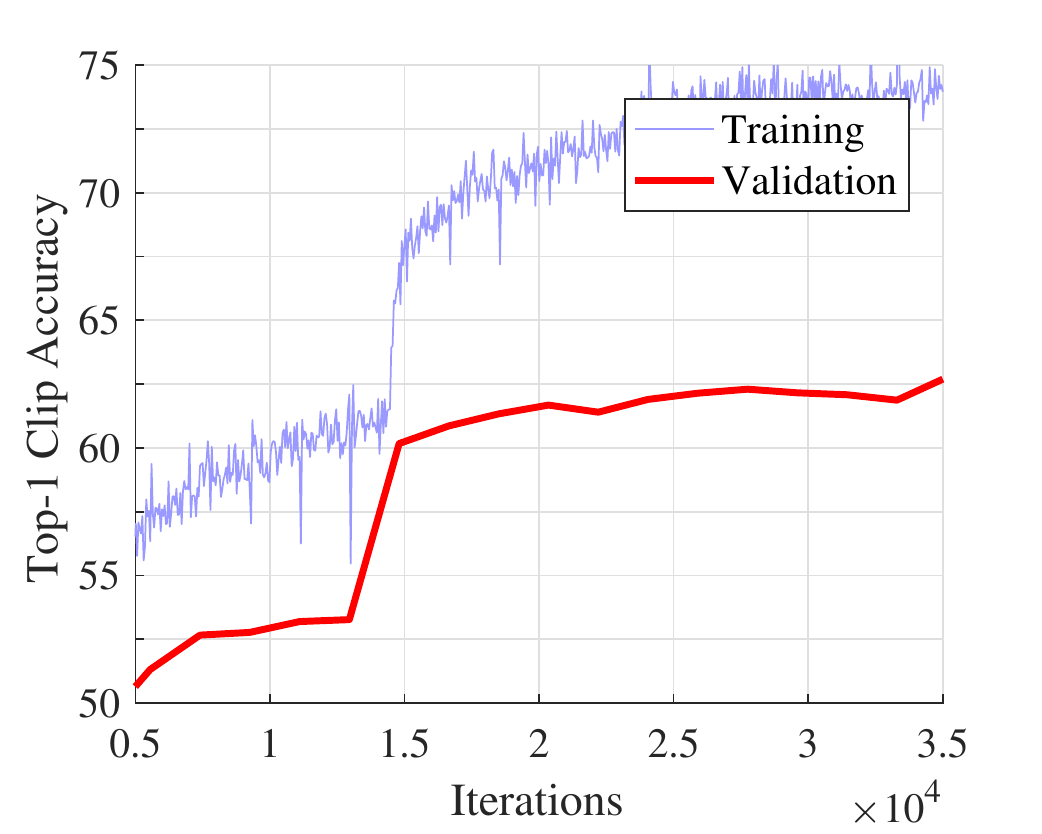}}}
\subfigure[]{\resizebox{0.4\textwidth}{!}{\includegraphics{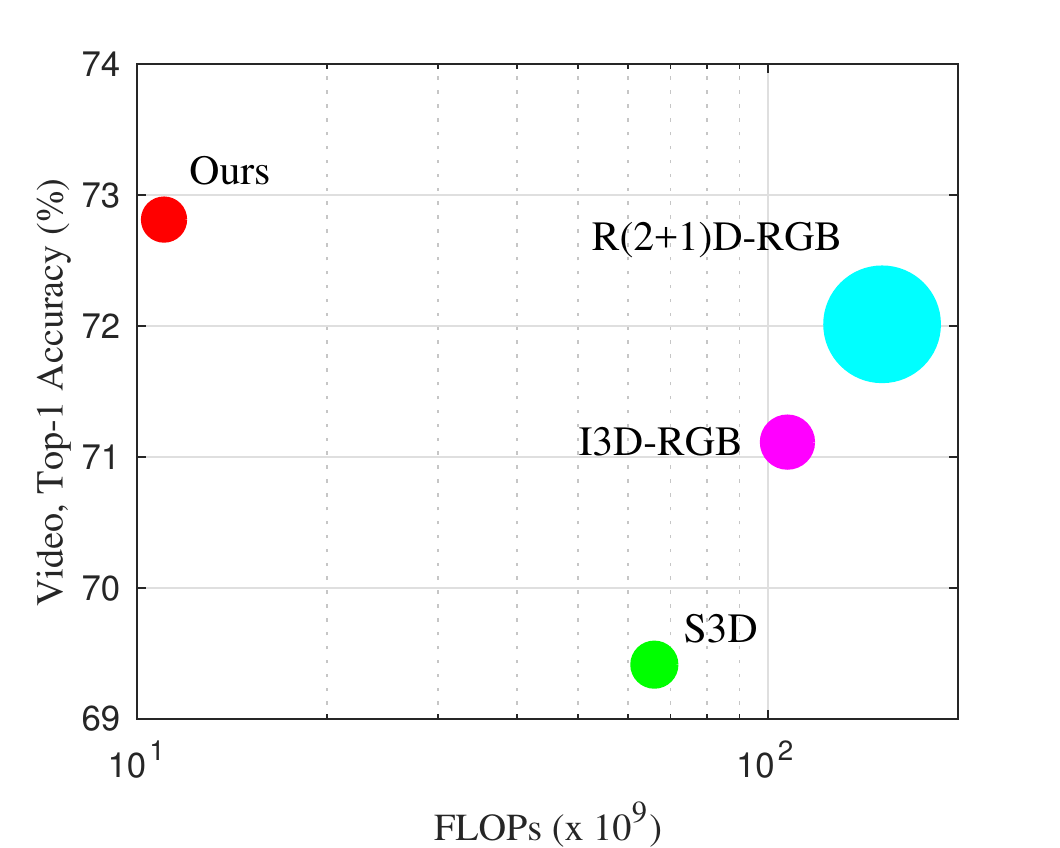}}}
\caption{Results on the Kinetics dataset (RGB Only). (a) The training and validation accuracy for \ourslong network. (b) Efficiency comparison between different 3D convolutional networks. The area of each circle is proportional to the total parameter number of the model.}
\label{fig_kinetics_curve_and_acc}
\end{figure}
In this experiment, the 3D MF-Net model is initialized by inheriting parameters from a 2D one (see Section~\ref{Network_Architecture}) pre-trained on the ImageNet-1k dataset. Then the 3D MF-Net is trained on Kinetics with an initial learning rate $0.1$ which decays step-wisely with a factor $0.1$. The weight decay is set to $0.0001$ and we use SGD as the optimizer with a batch size $1,024$. We train the model on a cluster of $64$ GPUs. Figure~\ref{fig_kinetics_curve_and_acc}(a) shows the training and validation accuracy curves, from which we can see the network converges fast and the total training process only takes about 36,000 iterations.

\begin{table}[t]
\caption{Comparison on action recognition accuracy with state-of-the-arts on Kinetics. The complexity is measured using FLOPs, \emph{i.e.} floating-point multiplication-adds. All results are only using RGB information, \emph{i.e.} no optical flow. Results with citation numbers are copied from the respective papers. }
\centering
\resizebox{1.\textwidth}{!}{
  \begin{tabular}{>{\centering}p{5cm}|>{\centering}p{2cm}|>{\centering}p{2cm}|>{\centering}p{2cm}|c}
  \toprule
  Method     							& \#Params 		& FLOPs  		& Top-1   		&~~~~~Top-5~~~~~ \\
  \hline
  Two-Stream~\cite{carreira2017quo}		&	12 M		& 	--			& 62.2 \%		&	--			\\
  ConvNet+LSTM~\cite{carreira2017quo}	& 	9 M			&	--			& 63.3 \%		&	--			\\
  \hline
  S3D~\cite{xie2017rethinking}       	&  8.8 M     	&  66.4 G  		& 69.4 \%  		& 89.1 \%  		\\
  I3D-RGB~\cite{carreira2017quo}     	& 12.1 M    	& 107.9 G 		& 71.1 \%  		& 89.3 \%  		\\
  R(2+1)D-RGB~\cite{tran2017closer} 	& 63.6 M    	& 152.4 G 		& 72.0 \%  		& 90.0 \%  		\\
  \midrule
  MF-Net (Ours)         				&\bf{8.0 M}    	& \bf{11.1 G} 	& \bf{72.8} \% 	& \bf{90.4} \%  \\
  \bottomrule
  \end{tabular}
}
\label{kinetics}
\end{table}
Table~\ref{kinetics} shows video action recognition results of different models trained on Kinetics. The models pre-trained on other large-scale video datasets, \emph{e.g.} Sports-1M~\cite{KarpathyCVPR14}, using substantially more training videos are excluded in the table for fair comparison. As can be seen from the results, 3D based CNN models significantly improve the Top-1 accuracy upon 2D CNN based models. This performance gap is because 2D CNNs extract features from each frame separately and thus are incapable of modeling complex motion features from a sequence of raw frames even when LSTM is used, which limits their performance. On the other hand, 3D CNNs can learn motion features end-to-end from raw frames and thus are able to capture effective spatio-temporal information for video classification tasks. However, these 3D CNNs are computationally expensive compared 2D ones.

In contrast, our proposed \ours is more computationally efficient than existing 3D CNNs. 
Even with a moderate number of fibers, the computational overhead introduced by the temporal dimension is effectively compensated and our \ourslong network only costs 11.1 GFLOPs, as low as regular 2D CNNs.
Regarding performance and parameter efficiency, our proposed model achieves the highest Top-1/Top-5 accuracy and meanwhile it has the smallest model size. Compared with the best $R(2+1)D$-$RGB$, our model is over $13 \times $ faster with $8 \times $ less parameters, yet achieving $0.8\%$ higher Top-1 accuracy. We note that the proposed model also costs the lowest GPU memory for both training and testing, benefiting from the optimized architecture mentioned in Section~\ref{Network_Architecture}.

\begin{figure}[t]	
	\center
	\resizebox{1.\textwidth}{!}{
		\includegraphics{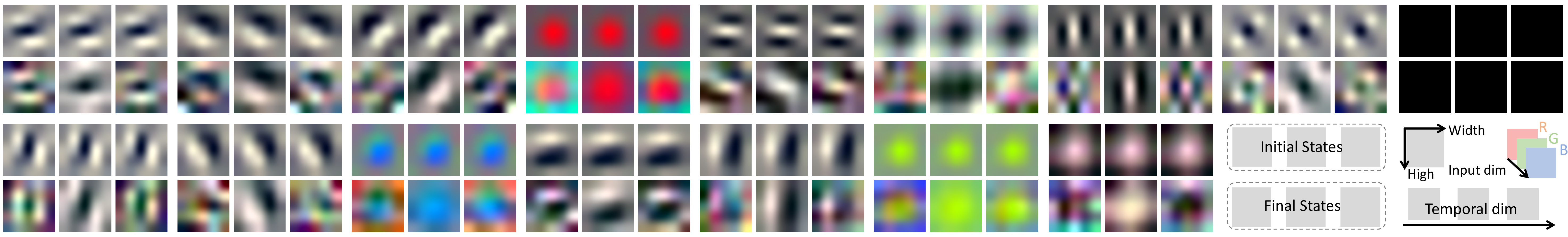}
	}
    \vspace{-0.2cm}
	\caption{Visualization of the learned filters. The filters initialized by the ImageNet pre-trained model using inflating are shown on the top. The corresponding learned 3D filters on Kinetics are shown at the bottom. (upscaled by 15x). Best viewed in color.
    }
	\label{fig_learnt_kernels}
    \vspace{-0.1cm}
\end{figure}

To get further insights into what our network learns, we visualize all 16 spatio-temporal kernels of the first convolutional layer in Figure~\ref{fig_learnt_kernels}. Each 2-by-3 block corresponds to two $3\times3\times5\times5$ filters, with the top and bottom rows showing the filter before and after learning, respectively. As the filters are initialized from a 2D network pretrained on ImageNet and inflated in the temporal dimension, all three sub-kernels are identical in the beginning. After learning, however, we see filters evolving along the temporal dimension with diverse patterns, indicating that spatio-temporal features are learned effectively and embedded in these 3D kernels.

\subsection{Video Classification with Fine-tuned Models}

In this experiment, we evaluate the generality and robustness of the proposed \ourslong network by transferring the features learned on Kinetics to other datasets. We are interested in examining whether the proposed model can learn robust video representations  that can generalize well to other datasets. We use the popular UCF-101~\cite{soomro2012ucf101} and HMDB51~\cite{kuehne2011hmdb} as  evaluation benchmarks. 

The UCF-101 contains $13,320$ videos from 101 categories and the HMDB51 contains $6,766$ videos from 51 categories. Both are divided into 3 splits. We follow experiment settings in~\cite{tran2015learning,tran2017convnet,tran2017closer,xie2017rethinking} and report the averaged three-fold cross validation accuracy. For model training on both datasets, we use an initial learning rate $0.005$ and decrease it for three times with a factor $0.1$. The weight decay is set to $0.0001$ and the momentum is set to $0.9$ during the SGD optimization. All models are fine-tuned using 8 GPUs with a batch size of 128 clips. 

\begin{table}[t]
\setlength{\tabcolsep}{10pt}
\centering
\caption{Action recognition accuracy on UCF-101 and HMDB51. The complexity is evaluated with FLOPs, \emph{i.e.} floating-point multiplication-adds. The top part of the table refers to related methods based on 2D convolutions, while the lower part to methods utilizing spatio-temporal convolutions. Column ``+OF'' denotes the use of Optical Flow.  FLOPs for computing optical flow are not considered.}
\resizebox{1.\textwidth}{!}{
  \begin{tabular}{>{\centering}p{4cm}|>{\centering}p{1.5cm}|>{\centering}p{1cm}|>{\centering}p{2cm}|c}
  \hline
  Method     										& 	FLOPs   	& +OF 				& UCF-101 	&~~~HMDB51~~~\\
  \hline
  ResNet-50~\cite{feichtenh2017spat} 				&   3.8 G     	& 			 		&  82.3 \% 	&  48.9 \% 	\\
  ResNet-152~\cite{feichtenh2017spat}				&   11.3 G     	& 			 		&  83.4 \% 	&  46.7 \% 	\\
CoViAR~\cite{wu2017compressed}	 					&   4.2 G     	& 			 		&  90.4 \% 	&  59.1 \% 	\\
  Two-Stream~\cite{simonyan2014two} 				&   3.3 G     	& \checkmark 		&  88.0 \% 	&  59.4 \% 	\\
  TSN~\cite{wang2016temporal}		   				&  	3.8 G     	& \checkmark		&  94.2 \% 	&  69.4 \%  \\
  \hline
  C3D~\cite{tran2015learning}        				&   38.5 G   	&        			&  82.3 \% 	&  51.6 \% 	\\
  Res3D~\cite{tran2017convnet}	   					& 	19.3 G   	&        			&  85.8 \% 	&  54.9 \% 	\\
  ARTNet~\cite{wang2017appearance}					&	25.7 G		&					&  94.3 \%  &  70.9 \%  \\
  
  I3D-RGB~\cite{carreira2017quo}	   				& 	107.9 G   	&        			&  95.6 \% 	&  74.8 \% 	\\
  R(2+1)D-RGB~\cite{tran2017closer}  				& 	152.4 G   	&        			&  96.8 \% 	&  74.5 \% 	\\
  \hline
  \ours (Ours)          							&  \bf{11.1 G} 	&        			&  96.0 \% 	&  74.6 \% 	\\
  \hline
  \end{tabular}
}
\label{finetune}
\end{table}

Table~\ref{finetune} shows results of the \ourslong network and comparison with state-of-the-art models. Consistent with above results, the \ourslong network achieves the state-of-the-art accuracy with much lower computation cost. In particular, on the UCF-101 dataset, the proposed model achieves $96.0\%$ Top-1 classification accuracy which is comparable with the sate-of-the-arts, but it is significantly  more computationally efficient  ($11.1$  vs. $152.4$ GFLOPs). Compared with Res3D~\cite{tran2017convnet} which is also based on ResNet backbone and costs about $19.3$ GFLOPs, the \ourslong network achieves over $10\%$ improvement in Top-1 accuracy ($96.0\%$ v.s. $85.8\%$) with $42\%$ less computational cost. 

\begin{figure}
\centering
\subfigure[UCF-101]{\resizebox{0.4\textwidth}{!}{\includegraphics{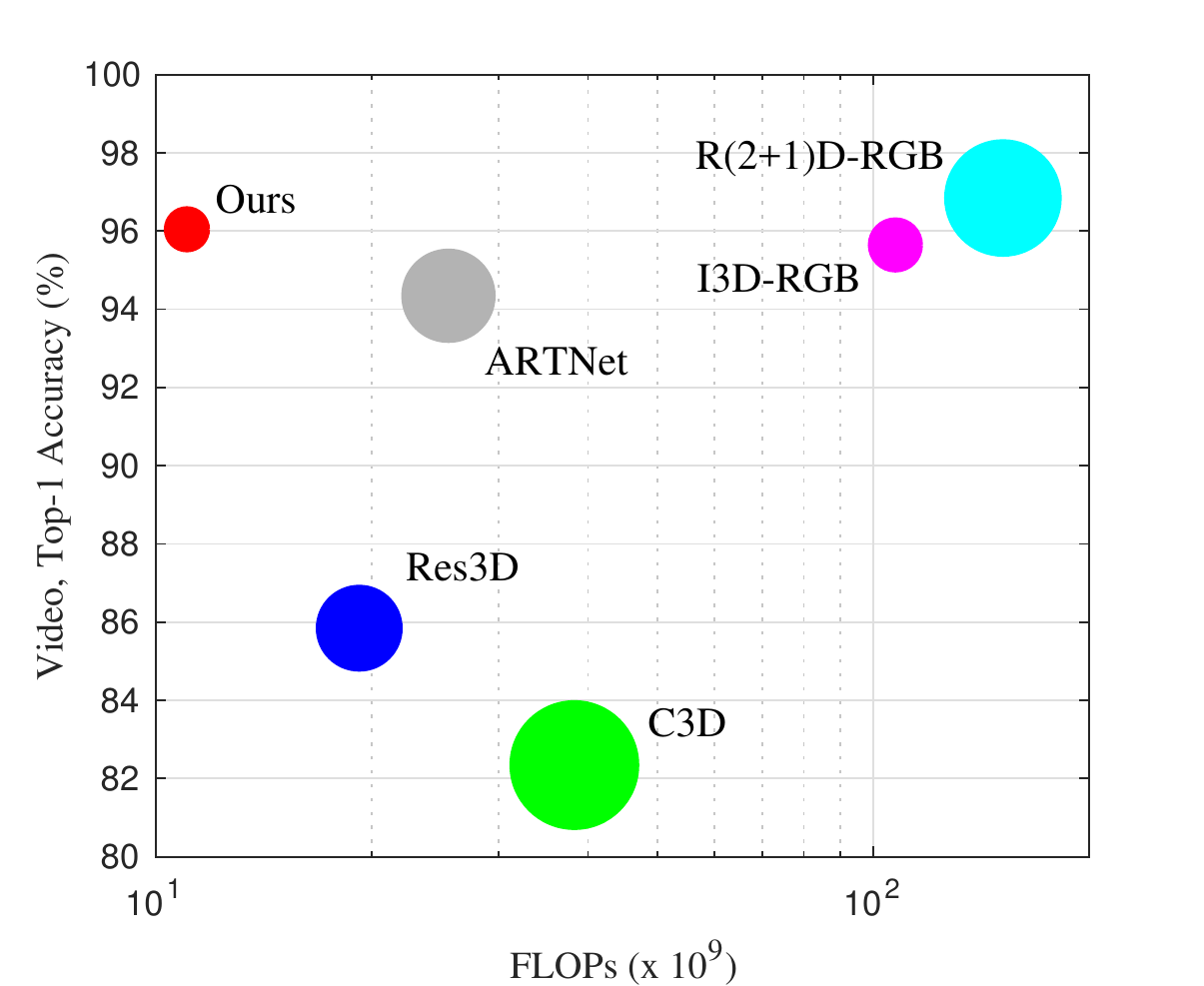}}}
\subfigure[HMDB51]{\resizebox{0.4\textwidth}{!}{\includegraphics{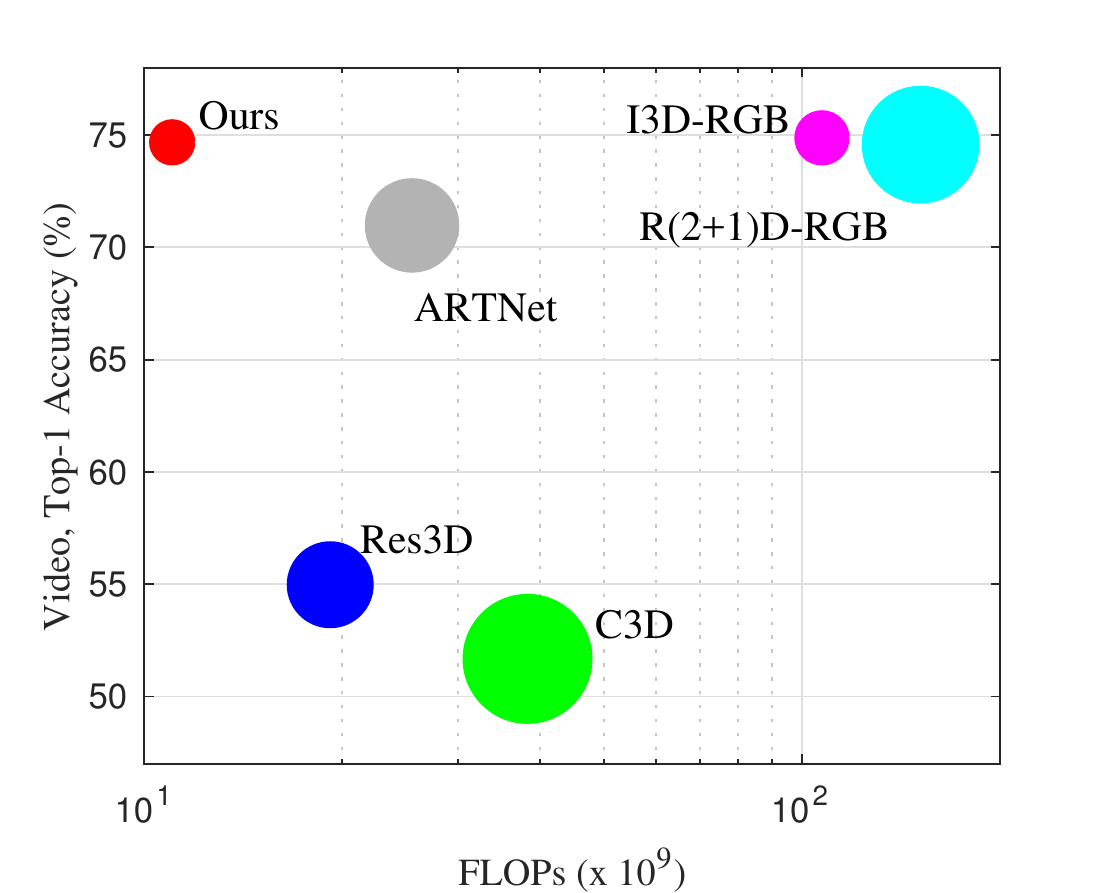}}}
\caption{Efficiency comparison between different methods. We use the area of each circle to show the total number of parameters for each model.}
\label{fig_ucf_hmdb}
\end{figure}
Meanwhile, the proposed \ourslong network also achieves the state-of-the-art accuracy on the HMDB51 dataset with significantly less computational cost. Compared with the 2D CNN based models that also only use RGB frames, our proposed model improves the accuracy by more than $15\%$ ($74.6\%$ v.s. $59.1\%$). Even compared with the methods that using extra optical information, our proposed model still improves the accuracy by over $5\%$. This advantage partially benefits from richer motion features that learned from large-scale video pre-training datasets, while 2D CNNs cannot. Figure~\ref{fig_ucf_hmdb} shows the results in details. It is clear that our model provides an order of magnitude higher efficiency than previous state-of-the-arts in terms of FLOPs but still enjoys the high accuracy.

\subsection{Discussion}
\label{sec:discussion}

The above experiments clearly demonstrate outstanding performance and efficiency of the proposed model. In this section, we discuss its potential limitations through success and failure case analysis on Kinetics.
\begin{figure}[h]
	\center
	\resizebox{0.85\textwidth}{!}{
		\includegraphics{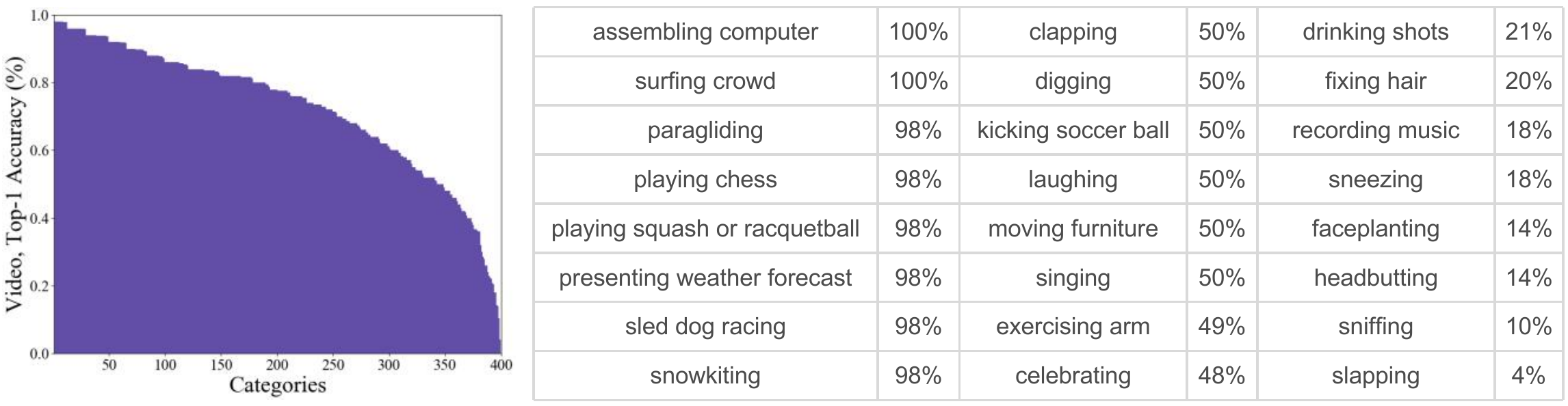}
	}
	\caption{Statistical results on Kinetics validation dataset. Left: Accuracy distribution of the proposed model on the validation set of Kinetics. The category is sorted by accuracy in a descending order. Right: Selected categories and their accuracy.}
	\label{fig_acc_dist}
\end{figure}

We first study category-wise recognition accuracy. We calculate the accuracy for each category and sort them in a descending order, shown in Figure \ref{fig_acc_dist} (left). Among all $400$ categories, we notice that $190$ categories have an accuracy higher than $80\%$ and $349$ categories have an accuracy higher than $50\%$. Only $17$ categories cannot be recognized well and have an accuracy lower than $30\%$. We list some examples along the spectrum in the right panel of Figure \ref{fig_acc_dist}. We find that in  categories with highest accuracy there are either some specific objects/backgrounds clearly distinguishable from other categories or specific actions spanning long duration. On the contrary, categories with low accuracy usually do not display any distinguishing object and the target action usually lasts for a very short time within a long video.

\begin{figure}[]
	\center
	\resizebox{0.95\textwidth}{!}{
		\includegraphics{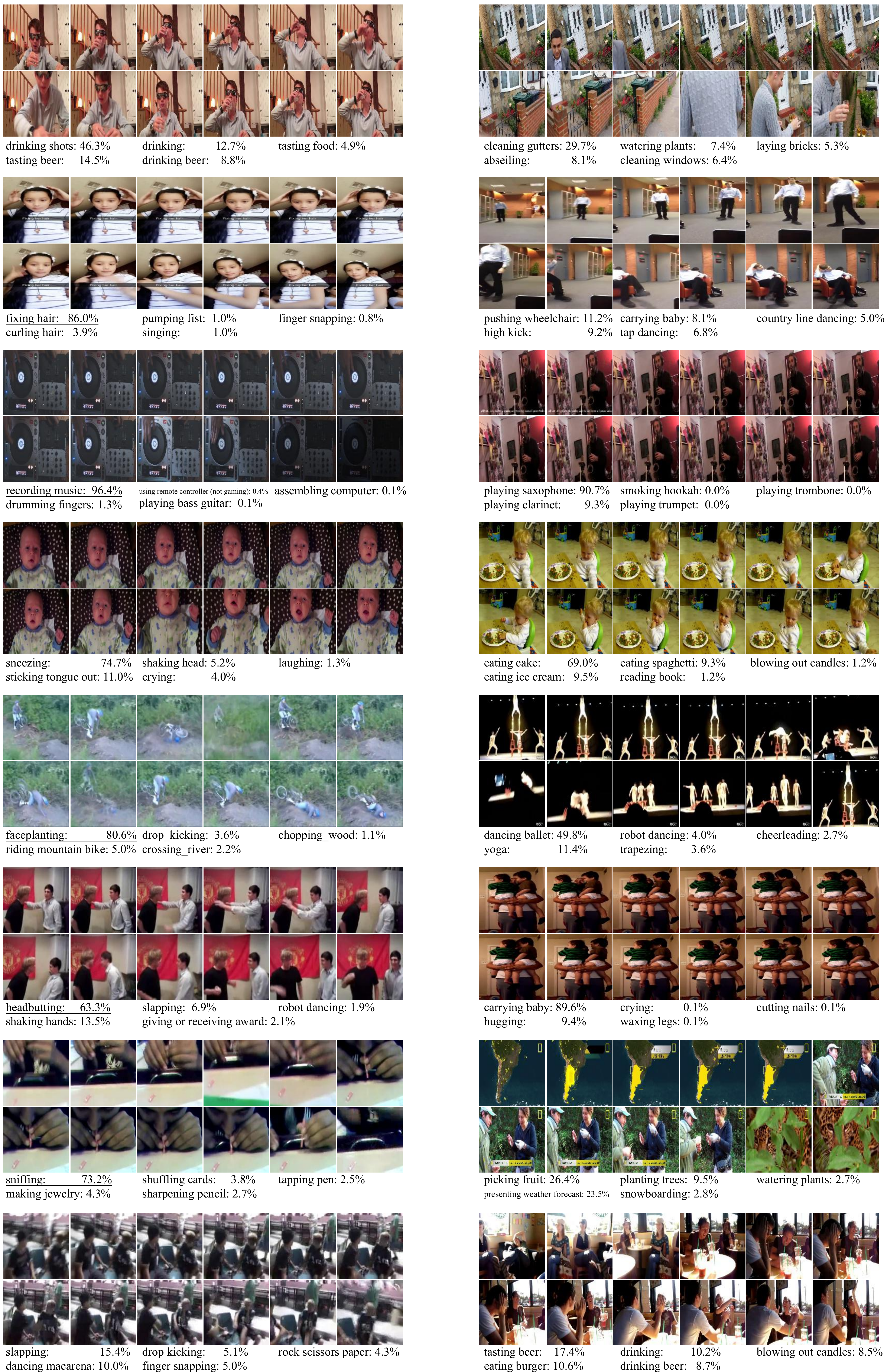}
	}
	\caption{Predictions made on the most difficult eight categories in Kinetics validation set. Left: Easy samples. Right: Hard samples. Top-5 confidence scores are shown below each video sequence. Underlines are used to emphasize correct prediction. Videos within the same row are from the same ground truth category.} 
	\label{fig_video_clips}
\end{figure}

To better understand success and failure cases, we visualize some of the video sequences in Figure~\ref{fig_video_clips}. The frames are evenly selected from the long video sequence. As can be seen from the results, the algorithm is more likely to make mistakes on videos without any distinguishable object or containing an action lasting a relatively short period of time.

\section{Conclusion}
\label{sec:conclusions}

In this work, we address the problem of building highly efficient 3D convolution neural networks for video recognition tasks. We proposed a novel \textit{multi-fiber} architecture, where sparse connections are introduced inside each residual block effectively reducing computations and a \switch is developed to compensate the information loss. Benefiting from these two novel architecture designs, the proposed model greatly reduces both model redundancy and computational cost. Compared with existing state-of-the-art 3D CNNs that usually consume an order of magnitude more computational resources than regular 2D CNNs, our proposed model costs significantly less resources yet achieves the state-of-the-art video recognition accuracy on Kinetics, UCF-101, HMDB51. We also showed that the proposed multi-fiber architecture is a generic method which can also benefit existing networks on image classification task.

\paragraph{Acknowledgements}
Jiashi Feng was partially supported by NUS IDS R-263-000-C67-646, ECRA R-263-000-C87-133 and MOE Tier-II R-263-000-D17-112.

\bibliographystyle{splncs}
\bibliography{reference}
\end{document}